# THE CORRECT AND EFFICIENT IMPLEMENTATION OF APPROPRIATENESS SPECIFICATIONS FOR TYPED FEATURE STRUCTURES

Dale Gerdemann and Paul John King*

Seminar für Sprachwissenschaft, Universität Tübingen[†]

## ABSTRACT

In this paper, we argue that type inferencing incorrectly implements appropriateness specifications for typed feature structures, promote a combination of type resolution and unfilling as a correct and efficient alternative, and consider the expressive limits of this alternative approach. Throughout, we use feature cooccurence restrictions as illustration and linguistic motivation.

## 1 INTRODUCTION

Unification formalisms may be either untyped (DCGs, PATR-II, LFG) or typed (HPSG). A major reason for adding types to a formalism is to express restrictions on feature cooccurences as in GPSG [6] in order to rule out nonexistant types of objects. For example, there are no verbs which have the feature +N. The simplest way to express such restrictions is by means of an appropriateness partial function Approp: Type × Feat → Type. With such an appropriateness specification many such restrictions may be expressed, though no restrictions involving reentrancies may be expressed.

In this paper, we will first in §2 survey the range of type constraints that may be expressed with just a type hierarchy and an appropriateness specification. Then in §3, we discuss how such type constraints may be maintained under unification as exemplified in the natural language parsing/generation system Troll [8].[1] Unlike previous systems such as ALE, Troll does not employ any type inferencing. Instead, a limited amount of named disjunction ([13], [5], [7]) is introduced to record type resolution possibilities. The amount of disjunction is also kept small by the technique of *unfilling* described in [10]. This strategy actually maintains appropriateness conditions in some cases in which a type inferencing strategy would fail. Finally, in §4, we discuss the possibilities for generalizing this approach to handle a broader range of constraints, including constraints involving reentrancies.

## 2 APPROPRIATENESS FORMALISMS

As discussed in Gerdemann & King [9], one can view appropriateness conditions as defining GPSG style feature cooccurence restrictions (FCRs). In [9], we divided FCRs into *conjunctive* and *disjunctive* classes. A conjunctive FCR is a constraint of the following form:

> if an object is of a certain kind
> then it deserves certain features
> with values of certain kinds

An FCR stating that a verb must have V and N features with values + and − respectively is an example of a conjunctive FCR. A disjunctive FCR is of the form:

---

*The research presented in this paper was partially sponsored by Teilprojekt B4 "Constraints on Grammar for Efficient Generation" of the Sonderforschungsbereich 340 of the Deutsche Forschungsgemeinschaft. We would also like to thank Thilo Götz for helpful comments on the ideas presented here. All mistakes are of course our own.

[†]Kl. Wilhelmstr. 113, D-72074 Tübingen, Germany, {dg,king}@sfs.nphil.uni-tuebingen.de.

[1] The Troll System was implemented in Quintus Prolog by Dale Gerdemann and Thilo Götz.

if an object is of a certain kind
then it deserves certain features
with values of certain kinds,
   or it deserves certain (perhaps
      other) features with values of
      certain (perhaps other) kinds,
   or ...
   or it deserves certain (perhaps
      other) features with values of
      certain (perhaps other) kinds

For example, the following FCR stating that inverted verbs must be auxiliaries is disjunctive: a verb must have the features INV and AUX with values + and +, − and +, or − and − respectively.

Both of these forms of FCRs may be expressed in a formalism employing finite partial order $\langle \mathsf{Type}, \sqsubseteq \rangle$ of types under subsumption, a finite set $\mathsf{Feat}$ of features, and an appropriateness partial function $\mathsf{Approp}: \mathsf{Type} \times \mathsf{Feat} \rightarrow \mathsf{Type}$. Intuitively, the types formalize the notion of *kinds of object*, $t \sqsubseteq t'$ iff each object of type $t'$ is also of type $t$, and $\mathsf{Approp}(t, f) = t'$ iff each object of type $t$ deserves feature $f$ with a value of type $t'$. We call such a formalism an *appropriateness formalism*. Carpenter's ALE and Gerdemann and Götz's Troll are examples of implementations of appropriateness formalisms.

How an appropriateness formalism encodes a conjunctive FCR is obvious, but how it encodes a disjunctive FCR is less so. An example illustrates best how it is done. Suppose that FCR $\rho$ states that objects of type $t$ deserve features $f$ and $g$, both with boolean values and furthermore that the values of $f$ and $g$ must agree. $\rho$ is the disjunctive FCR

   if an object is of type $t$
   then it deserves $f$ with value +
      and $g$ with value +,
      or it deserves $f$ with value −
      and $g$ with value −

To encode $\rho$, first introduce subtypes, $t'$ and $t''$ of $t$ ($t \sqsubseteq t', t''$), one subtype for each disjunct in the consequent of $\rho$. Then encode the feature/value conditions in the first disjunct by putting $\mathsf{Approp}(t', f) = +$ and $\mathsf{Approp}(t', g) = +$, and encode the feature/value conditions in the second disjunct by putting $\mathsf{Approp}(t'', f) = -$ and $\mathsf{Approp}(t'', g) = -$.[2]

This approach makes two important closed-world type assumptions about the types that subsume no other types (henceforth *species*). First, the *partition condition* states that for each type $t$, if an object is of type $t$ then the object is of exactly one species subsumed by $t$. Second, the *all-or-nothing condition* states that for each species $s$ and feature $f$, either every or no object of species $s$ deserves feature $f$.[3] An appropriateness formalism such as ALE ([2], [3]) that does not meet both conditions may not properly encode a disjunctive FCR. For example, consider disjunctive FCR $\rho$. An appropriateness formalism may not properly encode that $t'$ and $t''$ represent all and only the disjuncts in the consequent of $\rho$ without the partition condition. An appropriateness formalism may not properly encode the feature/value conditions demanded by each disjunct in the consequent of $\rho$ without the all-or-nothing condition.

As indicated above, ALE is an example of a formalism that does not meet both of these closed world assumptions. In ALE a feature structure is *well-typed* iff for each arc in the feature structure, if the source node is labelled with type $t$, the target node is labelled with type $t'$ and the arc is labelled with feature $f$ then $\mathsf{Approp}(t, f) \sqsubseteq t'$. Furthermore, a feature structure is *well-typable* iff the feature structure sub-

---

[2]This example FCR is, for expository purposes, quite simple. The problem of expressing FCR's, however, is a real linguistic problem. As noted by Copestake et al. [4], it was impossible to express even the simplest forms of FCRs in their extended version of ALE.

The basic principle of expressing FCRs also extends to FCRs involving longer paths. For example, to ensure that for the type $t$, the path $\langle fg \rangle$ takes a value subsumed by $s$, one must first introduce the chain $\mathsf{Approp}(t, f) = u$, $\mathsf{Approp}(u, g) = s$. Such intermediate types could be introduced as part of a compilation stage.

[3]Note that these closed world assumptions are explicitly made in Pollard & Sag (1994) [14].

sumes a well-typed feature structure. In ALE, type inferencing is employed to ensure that all feature structures are well-typable—in fact, all feature structures are well typed. Unfortunately, well-typability is not sufficient to ensure that disjunctive FCRs are satisfied. Consider, for example, our encoding of the disjunctive FCR $\rho$ and suppose that $\varphi$ is the feature structure $t[f:+,g:-]$. $\varphi$ is well-typed, and hence trivially well-typable. Unfortunately, $\varphi$ violates the encoded disjunctive FCR $\rho$.

By contrast, the Troll system described in this paper has an effective algorithm for deciding well-formedness, which is based on the idea of efficiently representing disjunctive possibilities within the feature structure. Call a well-typed feature structure in which all nodes are labelled with species a *resolved feature structure* and call a set of resolved feature structures that have the same underlying graph (that is, they differ only in their node labellings) a *disjunctive resolved feature structure*. We write $\mathcal{FS}$, $\mathcal{RFS}$ and $\mathcal{DRFS}$ for the collections of feature structures, resolved feature structures and disjunctive resolved feature structures respectively. Say that $F' \in \mathcal{RFS}$ is a *resolvant* of $F \in \mathcal{FS}$ iff $F$ and $F'$ have the same underlying graph and $F$ subsumes $F'$. Let type resolution be the total function $\mathcal{R}:\mathcal{FS} \to \mathcal{DRFS}$ such that $\mathcal{R}(F)$ is the set of all resolvants of $F$.

Guided by the partition and all-or-nothing conditions, King [12] has formulated a semantics of feature structures and developed a notion of a satisfiable feature structure such that $F \in \mathcal{FS}$ is satisfiable iff $\mathcal{R}(F) \neq \emptyset$. Gerdemann & King [9] have also shown that a feature structure meets all encoded FCRs iff the feature structure is satisfiable. The Troll system, which is based on this idea, effectively implements type resolution.

Why does type resolution succeed where type inferencing fails? Consider again the encoding of $\rho$ and the feature structure $\varphi$. Loosely speaking, the appropriateness specifications for type $t$ encode the part of $\rho$ that states that an object of type $t$ deserves features $f$ and $g$, both with boolean values. However, the appropriateness specifications for the speciate subtypes $t'$ and $t''$ of type $t$ encode the part of $\rho$ that states that these values must agree. Well-typability only considers species if forced to. In the case of $\varphi$, well-typability can be established by considering type $t$ alone, without the partition condition forcing one to find a well-typed species subsumed by $t$. Consequently, well-typability overlooks the part of $\rho$ exclusively encoded by the appropriateness specifications for $t'$ and $t''$. Type resolution, on the other hand, always considers species. Thus, type resolving $\varphi$ cannot overlook the part of $\rho$ exclusively encoded by the appropriateness specifications for $t'$ and $t''$.

## 3 MAINTAINING APPROPRIATENESS CONDITIONS

How may these $\mathcal{DRFS}$ be used in an implementation? A very important property of the class of $\mathcal{DRFS}$ is that they are closed under unification, i.e., if $F$ and $F' \in \mathcal{DRFS}$ then $F \sqcup F' \in \mathcal{DRFS}$.[4] Given this property, it would in principle be possible to use the disjunctive resolved feature structures in an implementation without any additional type inferencing procedure to maintain satisfiability. It would, of course, not be very efficient to work with such large disjunctions of feature structures. These disjunctions of feature structures, however, have a singular property: all of the disjuncts have the same shape. The disjuncts differ only in the types labeling the nodes. This property allows a disjunctive resolved feature structure to be represented more efficiently as a single untyped feature structure plus a set of dependent node labelings, which can be further compacted using named disjunction as in Gerdemann [7], Dörre & Eisele [5] or Maxwell & Kaplan [13].

---

[4]In fact, it can be shown that if $F$ and $F' \in \mathcal{FS}$ then $\mathcal{R}(F) \sqcup \mathcal{R}(F') = \mathcal{R}(F \sqcup F')$. Unification of sets of feature structures is defined here in the standard way: $S \sqcup S' = \{F \mid F' \in S$ and $F'' \in S'$ and $F = F' \sqcup F''\}$.

For example, suppose we type resolve the feature structure $t[f : bool, g : bool]$ using our encoding of $\rho$. One can easily see that this feature structure has only two resolvants, which can be collapsed into one feature structure with named disjunction as shown below:

$$\left\{ \begin{bmatrix} t' \\ f{:}+ \\ g{:}+ \end{bmatrix}, \begin{bmatrix} t'' \\ f{:}- \\ g{:}- \end{bmatrix} \right\} \Rightarrow \begin{bmatrix} \langle 1 \ t' \ t'' \rangle \\ f{:} \ \langle 1 \ + \ - \rangle \\ g{:} \ \langle 1 \ + \ - \rangle \end{bmatrix}$$

We now have a reasonably compact representation in which the FCR has been translated into a named disjunction. However, one should note that this disjunction is only present because the features $f$ and $g$ happen to be present. These features would need to be present if we were enforcing Carpenter's [2] total well typing requirement, which says that features that are allowed must be present. But total well typing is, in fact, incompatible with type resolution, since there may well be an infinite set of totally well typed resolvants of a feature structure. For example, an underspecified list structure could be resolved to a list of length 0, a list of length 1, etc.

Since total well typing is not required, we may as well actively unfill redundant features.[5] In this example, if the $f$ and $g$ features are removed, we are left with the simple disjunction $\{t', t''\}$, which is equivalent to the ordinary type $t$.[6] Thus, in this case, no disjunction at all is required to enforce the FCR. All that is required is the assumption that $t$ will only be extended by unifying it with another (compacted) member of $\mathcal{DRFS}$.

---

[5]Intuitively, features are redundant if their values are entirely predictable from the appropriateness specification. See Götz [10], Gerdemann [8] for a more precise formulation.

[6]In this case, it would also have been possible to unfill the original feature structure before resolving. Unfortunately, however, this is not always the case, as can be seen in the following example: $t[f : +] \Rightarrow \{t'[f : +]\} \Rightarrow t'$.

This, however, was a simple case in which all of the named disjunction could be removed. It would not have been possible to remove the features $f$ and $g$ if these features had been involved in reentrancies or if these features had had complex values. In general, however, our experience has been that even with very complex type hierarchies and feature structures for HPSG, very few named disjunctions are introduced.[7] Thus, unification is generally no more expensive than unification with untyped feature structures.

## 4 CONCLUSIONS

We have shown in this paper that the kind of constraints expressible by appropriateness conditions can be implemented in a practical system employing typed feature structures and unification as the primary operation on feature structures. But what of more complex type constraints involving reentrancies? Introducing reentrancies into constraints allows for the possibility of defining recursive types, such as the definition of append in [1]. Clearly the resolvants of such a recursive type could not be precompiled as required in Troll.

One might, nevertheless, consider allowing reentrancy-constraints on non-recursively defined types. A problem still arises; namely, if the resolvants of a feature structure included some with a particular reentrancy and some without, then the condition that all resolvants have the same shape would no longer hold. One would therefore need to employ a more complex version of named disjunction ([13], [5], [11]). It is questionable whether such additional complexity would be justified to handle this limited class of reentrancy-constraints.

It seems then, that the class of constraints that can be expressed by appro-priateness conditions corresponds closely to the class of constraints that can be efficiently precompiled. We take this as a justification for appropriateness formalisms in general. It makes sense to abstract out the efficiently processable constraints and then allow another mechanism, such as attachments of definite clauses, to express more complex constraints.

---

[7]Our experience is derived primarily from testing the Troll system on a rather large grammar for German partial verb phrases, which was written by Erhard Hinrichs and Tsuneko Nakazawa and implemented by Detmar Meurers.